\documentclass[twoside,leqno,twocolumn]{article}

\usepackage[letterpaper]{geometry}

\usepackage{ltexpprt}
\usepackage{hyperref}

\usepackage{algorithmic}  
\usepackage{algorithm} 
\usepackage[algo2e]{algorithm2e} 
\usepackage{pifont,dsfont}
\usepackage{makecell}
\usepackage{amsfonts}
\usepackage{wrapfig,lipsum,booktabs}
\usepackage[export]{adjustbox}
\usepackage{xcolor,colortbl}

\usepackage{graphicx}
\usepackage{subfigure}

\usepackage{refcount}
\usepackage{amsfonts}
\usepackage{caption}
\usepackage{makecell}

\newcommand{\myModel}{\textsc{CoRnNet}}

\newtheorem{pro-stat}{Problem Definition}

\newcommand{\bh}[1]{{\small\color{violet}{\bf BH: #1}}}
\newcommand{\hide}[1]{}

\usepackage[most]{tcolorbox}
\definecolor{block-gray}{gray}{0.85}
\newtcolorbox{myquote}{colframe=black,boxrule=1pt,
colback=white,grow to right by=10mm,grow to left by=8mm,
boxsep=0pt,breakable}

\newtcolorbox{inside_myquote}{boxrule=0pt,
colback=block-gray,grow to right by=3mm,grow to left by=3mm,
top=0pt,bottom=0pt}

\begin{document}

\newcommand\relatedversion{}
\renewcommand\relatedversion{\thanks{The full version of the paper can be accessed at \protect\url{https://arxiv.org/abs/1902.09310}}} 

\title{\Large Conversational Question Answering with Reformulations over Knowledge Graph}
\author{Lihui Liu\thanks{University of Illinois at Urbana-Champaign, \{lihuil2, blaineh2, htong\}@illinois.edu}
\and Blaine Hill$^*$
\and Boxin Du\thanks{Amazon, \{boxin, feiww\}@amazon.com}
\and Fei Wang$^\dagger$
\and Hanghang Tong$^*$}

\date{}

\maketitle


\fancyfoot[R]{\scriptsize{Copyright \textcopyright\ 2024 by SIAM\\
Unauthorized reproduction of this article is prohibited}}





\begin{abstract} \small\baselineskip=9pt

Conversational question answering (ConvQA) over knowledge graphs (KGs) involves answering multi-turn natural language questions about information contained in a KG. State-of-the-art methods of ConvQA often struggle with inexplicit question-answer pairs. These inputs are easy for human beings to understand given a conversation history, but hard for a machine to interpret, which can degrade ConvQA performance. To address this problem, we propose a reinforcement learning (RL) based model, \myModel, which utilizes question reformulations generated by large language models (LLMs) to improve ConvQA performance. 
\myModel\ adopts a teacher-student architecture where a teacher model learns question representations using human writing reformulations, and a student model to mimic the teacher model’s output via reformulations generated by LLMs. 
The learned question representation is then used by a RL model to locate the correct answer in a KG. 
Extensive experimental results show that \myModel\ outperforms state-of-the-art ConvQA models.

\end{abstract}

\textbf{Keywords:} Knowledge graph conversational question answering; Reinforcement learning

\section{Introduction}

Knowledge graphs (KGs) are collections of nouns represented as nodes (representing real-world entities, events, and objects) and edges (denoting relationships between nodes). Knowledge graph question answering (KGQA) has long been a focus of study, with the goal of answering queries using information from a KG. However, traditional KGQA approaches often only consider single-shot questions, rather than the iterative nature of real-world conversation. Conversational question answering (ConvQA) addresses this gap by allowing users to interact with a QA system conversationally. ConvQA systems have had much success, as seen by Apple's Siri, Amazon's Alexa and OpenAI's ChatGPT.

Conversational question answering (ConvQA) involves a multiturn process consisting of users iteratively asking natural language questions, a system deciphering both the conversation context and underlying queries, and the system returning natural language answers. 
Some models will create rich, human-like responses ~\cite{alexa, xiaobing, gpt3}, these methods are known as `dialogue' conversation models. While for ConvQA over KGs, a corresponding entity in the KG is sufficient to answer the input question, we call it `non-dialogue' conversation models.
In this paper, we focus on the non-dialogue ConvQA task as shown in Example 1.

\hide{
Once the system obtains an answer embedding, it is sufficient to map this to the corresponding node in the KG \bh{maybe include an example of dialogue vs non-dialogue answer?} and forgo the need to train a natural language decoder ~\cite{ask_the_right_question, qa_rewrite} to create rich, human-like responses ~\cite{alexa, xiaobing, gpt3} for the benefit of the user: ConvQA models which include this decoder are known as 'dialogue' models, while models that do not are known as 'non-dialogue' models. 
In this paper, the authors focus on the non-dialogue ConvQA task as shown in Example 1.
}
\scalebox{0.8}{
\begin{myquote}

Example 1:

$q_1$: Who is the author that wrote the book Moby-Dick?
\begin{inside_myquote}
Reformulation1: Author of the book?

Reformulation2: Who wrote Moby Dick?
\end{inside_myquote}
$a^1$: Herman Melville

$q_2$: When was he born?
\begin{inside_myquote}
Reformulation1: His birthdate is?

Reformulation2: When was Herman Melville born?
\end{inside_myquote}
$a^2$: 1 August 1819

$q_3$: And where is he from originally?
\begin{inside_myquote}
Reformulation1: His place of birth?

Reformulation2: Where did he grow up?
\end{inside_myquote}
$a^3$: Manhattan

$q_4$:How about his wife?
\begin{inside_myquote}
Reformulation1: Where is Herman Melville's wife from?

Reformulation2: Herman Melville's wife's place of birth?
\end{inside_myquote}
$a^4$: Boston

$q_5$: Did they make a movie based on the book?

$a^5$: yes

\end{myquote}
}

In general, a conversation is typically initiated with a well-formed  question (i.e., $q_1$) followed by inexplicit follow-up questions (e.g., $q_2$ - $q_5$). The initial question ($q_1$) often includes a central \textbf{topic entity} of interest ("Moby-Dick"), while the topic entities of follow-up questions ($q_2$ - $q_5$) are not explicitly given. Additionally, the topic entity of the conversation may shift over time (e.g., inquiring about the birth time of Herman Melville in $q_2$). 

To operate ConvQA over KG, different methods have previously been proposed. For instance, Magdalena et al. in ~\cite{conquer} use named entity recognition (NER) methods to detect potential KG topic entities in the conversation and employ multi-agent reinforcement learning starting from these entities to find answers; the performance of this method is largely dependent on the quality of the detected entities. Philipp et al. in ~\cite{convex} propose finding a conversation-related subgraph and using heuristic-based methods to identify the answer within the subgraph. The subgraph is expanded as new questions are asked. Endri et al. in ~\cite{contrastiveQA} use contrastive learning to make KG entity embeddings dissimilar from one another, enabling the model to separate correct answers from incorrect answers. 

Despite the above achievements, inexplicit input data hinders a ConvQA system's ability to find correct answers ~\cite{conquer, qa_rewrite, ask_the_right_question}. 
Two common linguistic phenomena which undermine the semantic completeness of a query in the conversation are: \textbf{anaphora} and \textbf{ellipsis} ~\cite{qa_rewrite}. 
\textbf{Anaphora} refers to the phenomenon of an expression that depends on an expression in the previous context. For example, in Example 1, the word ``he" in $q_2$ refers to $a^1$. 
Meanwhile, \textbf{ellipsis} refers to the phenomenon of the omission of expressions in the previous context. For example, the complete form $q_4$ should be "Where is Herman Melville's wife from?". 
To address this issue, several methods aim to learn a \textbf{reformulation} of the input query, \textit{rewriting the original question} in a more meaningful way. Then, one can search for the answer using this new reformulation with existing techniques ~\cite{ask_the_right_question, qa_rewrite, SL-Oracle}. 
Despite many of the existing question rewriting models have shown potential to enhance ConvQA performance, as demonstrated by prior research~\cite{reformulation_not_good}, their generated reformulations fall short compared to human-generated reformulations ~\cite{qa_rewrite}.

\begin{table}[]
	\centering
	\caption{Notation and definitions}
 \vspace{-1\baselineskip}
	\footnotesize
	\begin{tabular}{|c|c|}
		\hline
		{\bf Symbols}       & {\bf Definition}                \\ \hline
		$\mathcal{G}=(\mathcal{V}, \mathcal{R}, \mathcal{L})$ & the knowledge graph \\ 
		$v_i$         & the $i^\textrm{th}$ entity/node in knowledge graph \\ 
		$r_i$         & the $i^\textrm{th}$ relation/edge in knowledge graph \\ 
		$\mathbf{e}_{v_i}$ / $\mathbf{e}_{i}$        & the embedding of node $v_i$ \\
		$\mathbf{r}_{i}$         & the embedding of relation $r_i$ \\
        $\mathbf{u}_{r_i}$ & the unique embedding of edge $r_i$ \\ \hline
		$C$           & the conversation \\ 
		$T$           & the number of turns in $C$ \\ 
		$q_i$         & the ith question in $C$ \\ 
        $\textrm{Ref}_{q_i}^k$ & the $k$th reformulation of question $q_i$ \\ 
		$a^i$         & the answer of question $q_i$ \\ 
        $w_i^t$  & the $i$th word in question $q_t$ \\
		$v_{q_1}$ & the main topic entity of the conversation \\ \hline
        $s \in S$ & RL states \\
        $A_s$ & the set of possible actions at state s \\
        $\mathbf{A}_s$ & the corresponding matrix of $A_s$ \\
        $a_i \in A_s$ & actions $a_i$ at state s \\
        $n_i$ & the entity in state $s_i$ \\
        $R$ & reward \\
        $\theta$ & parameters of policy network \\
        $\pi_{\theta}$  &  policy parameterized by $\theta$  \\ \hline
	\end{tabular}
\label{notation}
\vspace{-2\baselineskip}
\end{table}
In this paper, we present \textbf{\myModel}, a new reinforcement learning (RL) model for non-dialogue conversational question answering (ConvQA) with large language model (LLM) generated reformulations. 
First, we fine-tune existing LLMs, GPT2 ~\cite{gpt2} and Bart ~\cite{bart}, to generated high quality reformulations, using human writing reformulations as the ground truth. Second, to further increase the convQA performance,
we propose a teacher-student architecture to achieve near human-level performance \footnote{Human-level performance refers to the ability to find answers based on real human writing reformulations during testing.}.
Specifically, \myModel\ (1) \textbf{directly trains} a teacher model with human writing reformulations in the training data, and (2) \textbf{indirectly trains} a student model with LLMs generated reformations to mimic the teacher model's output so that it can approach human-level performance. Note that the human writing reformulations only exist in the training and validation data.
\hide{
Between turns and prior to the start of QA, identifying the topic entity is a necessity. \myModel\ performs this by examining any previous current topic entities and (re)evaluates based on a feed forward neural network (NN) classifier. If \myModel\ determines that the topic entity has likely changed, it selects the topic entity via a second NN.
}
Lastly, to locate an answer, a RL model walks over the KG, sampling actions from a policy network to guide the direction of the walk and identify candidate answers.
Our experiments demonstrate the effectiveness of \myModel\ and its superiority over the state-of-the-art conversational question answering baselines.

The main contributions of this paper are:
\begin{itemize}
  \item {\bf Analysis.} We demonstrate that although LLMs are good question reformulators, their performance lags behind human-level performance.
  \item {\bf Algorithm}. We propose a RL based model \myModel\ which utilizes the question reformulations to improve the QA performance. The proposed teacher-student model can help us achieve near human-level performance with LLMs generated reformulations. 
  \item {\bf Empirical Evaluations}. The experimental results on several real-world datasets demonstrate that the proposed \myModel\ consistently achieves state-of-the-art performance.
\end{itemize}


\section{Problem Definition}\label{problem-definition}

Table ~\ref{notation} gives the main notation used throughout this paper. 
Uppercase letters are used for matrices, sets or constant value (e.g., $C, T$). Bold lowercase letters are for vectors or embedding (e.g., $\mathbf{r}_{i}$) and lowercase letters (e.g., $s, a_i$) for scalars or variables.
A KG can be denoted as $\mathcal{G}=(\mathcal{V}, \mathcal{R}, \mathcal{L})$ where $\mathcal{V} = \{v_1, v_2, ..., v_n\}$ is the set of nodes/entities, $\mathcal{R} = \{r_1, r_2, ..., r_m\}$ is the set of relations and $\mathcal{L}$ is the list of triples.
Each triple in the KG can be denoted as $(h, r, t)$ where $h \in \mathcal{V}$ is the head (i.e., subject) of the triple, $t \in \mathcal{V}$ is the tail (i.e., object) of the triple and $r \in \mathcal{R}$ is the edge (i.e., relation, predicate) of the triple which connects the head $h$ to the tail $t$. 
The embedding of a node or relation type is represented by bold lowercase letters, e.g., $\mathbf{e}_{i}$, $\mathbf{r}_{i}$. Each triple/edge $(h, r, t)$ in the KG has a unique edge embedding which is denoted as $\mathbf{u}_{r}$.

Conversational question answering over a KG aims to iteratively answer multiple related questions from the users. 
Unlike dialog question answering which wants the chatbot to imitate the response of a human, ConvQA over KG only requires the model to return entities in the knowledge graph. We formally define the key terminologies used in this paper as follows. 

\begin{figure*}
	\centering
	\includegraphics[width=0.99\textwidth]{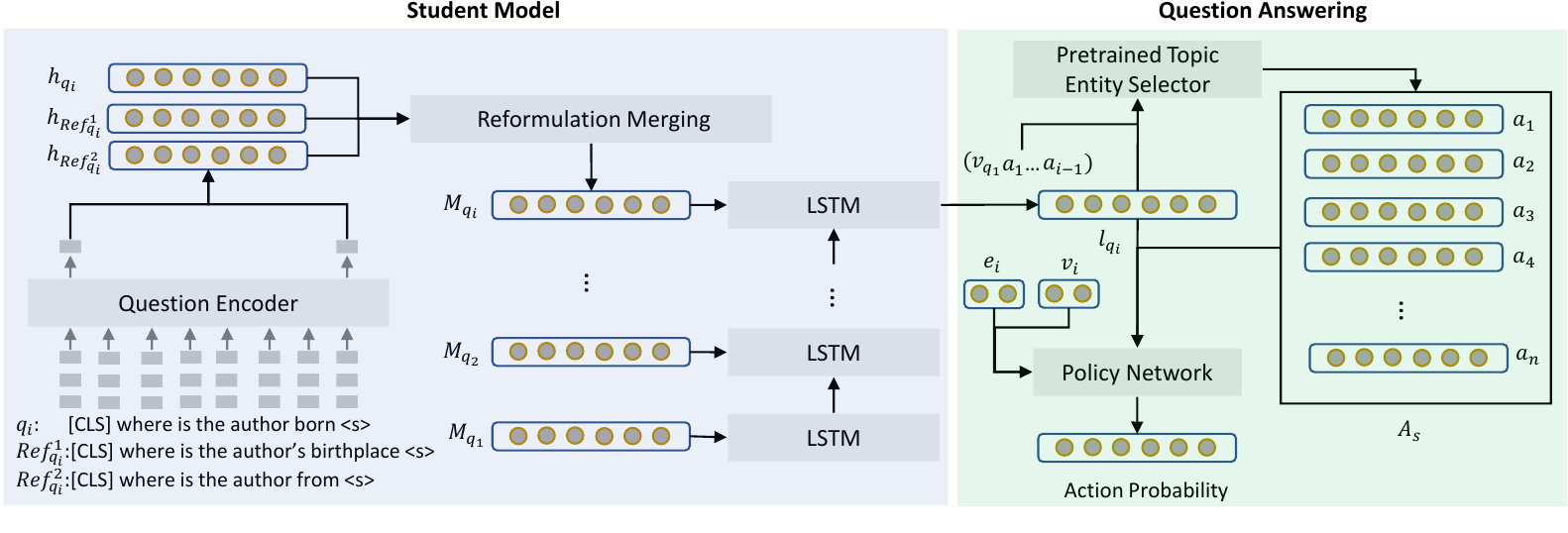}
 \vspace{-1\baselineskip}
	\caption{Training Process of \myModel. The light gray part denotes the architecture of the student model. The light green part shows the framework of RL-based question answering model. Both are trained end-to-end.
 }
    \label{framework}
    \vspace{-1\baselineskip}
\end{figure*}

\noindent \textbf{Conversation}.
A conversation $C$ with $T$ turns is made up of a sequence of questions 
{$q_1, q_2, ..., q_T$} and their corresponding answers Ans = \{ $a^1, a^2, ..., a^T$ \}, such that $C$ = ⟨($q_1$, $a^1$), ($q_2$, $a^2$), ..., ($q_T$ , $a^T$ )⟩. Example 1 in Introduction contains $T$ = 5 turns. 
We assume that $q_1$ is well-formed, and all other $q_t$ are inexplicit.

\noindent \textbf{Question}. Each question $q_t$ is a sequence of words $q_t = (w_1^t, . . . , w_{	\Omega_t}^t)$, where $\Omega_t$ is the number of words in $q_t$. 
We assume that each question can be mapped to a unique relation $r_{q_t}$ in the KG and make no assumptions on the grammatical correctness of $q_t$. 

\noindent \textbf{Topic Entity}. We assume that each $q_t$ has a topic/central entity $v_{q_t}$ which the user wants to ask about.
We assume that the topic entity of $q_1$ is given in the training data, while the topic entities for other questions $q_2, ..., q_T$ are not given. 
For example, for the five questions in Example 1, their topic entities are {\tt Moby Dick}, {\tt Herman Melville}, {\tt Herman Melville}, {\tt Moby Dick} and {\tt Moby Dick}, respectively. 
The topic entity of $q_1$ is presumed the main topic entity which is denoted as $v_{q_1}$. 

\noindent \textbf{Answer}. Each answer $a^t$ to question $q_t$ is a (possibly multiple, single, or null-valued) set of entities in the KG. We assume that all the answer entities exist in the KG, except true or false questions. 

\noindent \textbf{Reformulation}.  
A reformulation is a sentence which expresses the same information as the input question, but in a different way. We assume in the training data, each question has multiple reformulations. 

\noindent \textbf{Turn}. Each question in $C$, including its reformulations and corresponding answers, constitutes a turn $t_i$. Each turn $t_i$ contains a question $q_i$, the answer $a^i$ and reformulations of $q_i$.

Based on the above, we formally define the problem of ConvQA over KG as: 
\begin{pro-stat}{Conversational Quesion Answering over Knowledge Graph}\label{pro_def}
	
	\textbf{Given:} (1) A knowledge graph G, (2) the training set of conversations where each question contains multiple human writing reformulations, (3) the test set of conversations where no question reformulation is provided;
	
	\textbf{Output:} (1) The trained model, (2) the answer for each question in each conversation of the test set. 
	
\end{pro-stat}

\subsection{Preliminaries: Reinforcement Learning}


The RL problem can usually be formulated as Markov Decision Processes (MDPs).
An MDP is defined by $M = (S, A, R, P, \gamma)$,
where $S$ is the state space and $A$ is the action space.
$R: S \times A \longrightarrow R$ is the reward function from the environment which maps a state-action pair to a scalar which denotes how much reward the agent can receive, 
and $P: S \times A \longrightarrow S$ is the transition function which defines the probability of transiting from a state-action pair to the next state. $\gamma$ is a discount factor where $\gamma \in [0, 1]$.
When modeling the KGQA problem as an RL task, the agent will learn a policy to find the correct answer in the KG, and make decisions guided by the policy function when it receives new queries.


\section{Proposed Method}\label{overview}

Due to anaphora and ellipsis, current ConvQA methods often rewrite input queries to generate more understandable reformulations. 
In this paper, we follow this idea by fine-tuning two existing LLMs, GPT2 and Bart, to generate reformulations. 
Despite GPT2 and Bart are good reformulation generators, their performance still lags behind human-level performance.
To further improve the performance, we propose a teacher-student architecture. 
The teacher model learns the question representation by using human writing reformulations, while the student model takes reformulations generated by LLMs as input, and tries to mimic the output of the teacher model, so that it can achieve the same performance as the teacher model despite using the LLMs generated reformulations. 

In each iteration, our model uses the conversation history and the current query to identify the current topic entity, and an RL agent travels the KG starting from the topic entity to find the answer. This process is repeated for a  number of turns until the conversation is completed. Figure ~\ref{framework} illustrates the framework of the proposed \myModel. We will describe the details of each component in the following subsections.                                                                         

\subsection{Student Model: LLMs Reformulation Encoder.}\label{qa_encoder_decoder}

\hide{
The architecture is show in Figure ~\ref{encoder_decoder}.

\begin{figure}
	\centering
	\includegraphics[width=0.43\textwidth]{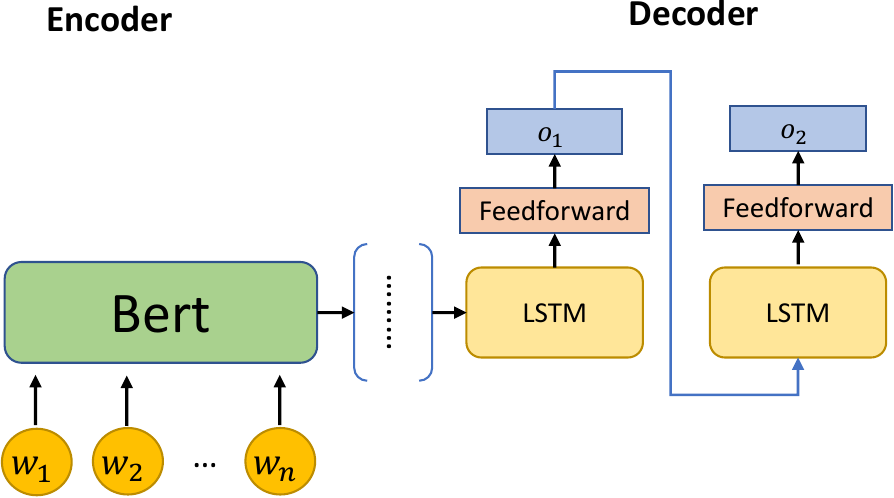}
	\caption{Question Encoder Decoder.}
	\label{encoder_decoder}
\end{figure}
}


\noindent{\bf A - Context Encoder.} 
Given a question $q_i = (w_1^i, w_2^i, . . . , w_{\Omega_t}^i)$, we first add two indicator tokens ([CLS] and $<$s$>$) to the beginning and end of the question context to signify its boundary. Then, we pass the processed question context through a pre-trained BERT ~\cite{bert} to extract contextual embeddings for each token: 
\begin{equation}
[\mathbf{h}_{CLS}, \mathbf{w}_1, ..., \mathbf{w}_{\Omega_t}, \mathbf{h}_{\mathbf{s}}] = \textrm{BERT}([CLS], w_1, . . . , w_{\Omega_t}, <s>)
\end{equation}
\noindent where $\mathbf{h}_{CLS}$ is the embedding of the [CLS] token and $\mathbf{h}_{\mathbf{s}}$ is the embedding of the $<$s$>$ token. The context question embedding is obtained from the transformation of $\mathbf{h}_{CLS}$ and  $\mathbf{h}_{\mathbf{s}}$, where $\textrm{FFN}$ is a feed forward neural network.
\begin{equation}
\mathbf{h}_{q_i} = \textrm{FFN}(\mathbf{h}_{CLS} || \mathbf{h}_{\mathbf{s}})
\end{equation}

\noindent{\bf B - Context Fusion.}
During training, each input question has multiple corresponding reformulations generated. For each reformulation, we use the Context Encoder to obtain its context embedding. To merge the reformulation information, we stack the embeddings of the $N$ reformulations and the original question context embedding to create a sequence with N+1 embeddings. 
We treat this sequence as the embedding of a language sentence and pass it through a Transformer Encoder ~\cite{transformer} to merge them together.
\begin{align*}
\mathbf{M_{q_i}} = \textrm{TRANSFORMER}([\mathbf{h}_{q_i} | \mathbf{h}_{{Ref}_{q_i}^{1}} | ... | \mathbf{h}_{{Ref}_{q_i}^{n}}])[0]
\end{align*}
where $\mathbf{M_{q_i}}$ is the query embedding after merging the reformulations.

\noindent{\bf C - Integrating Conversational History.}
Another problem in ConvQA is that the user’s inputs are often ambiguous, hampering a system's ability to give accurate answers. This is illustrated in Example 1 $q_3$ ``And where is he from originally?". It is impossible to identify the antecedent to the pronoun `he' without any conversational history. Consequently, \textit{conversational history is vital} to the success of \myModel. We use an LSTM to encode all the conversational history which is given below. 
\begin{equation}
\mathbf{{l}_{q_i}} = \textrm{LSTM}(\mathbf{M_{q_i}})
\end{equation}
the output of the LSTM $\mathbf{{l}_{q_i}}$ will be treated as the query embedding and be used by other components. Note that the reformulations used here are generated by LLMs. 

\subsection{Teacher Model: Human Writing Reformulation Encoder}

Reformulations have been used by various methods to improve the performance of QA systems by creating more understandable queries. For instance, in ~\cite{ask_the_right_question}, Christian et al. use a Seq2Seq-based reinforcement learning agent to transform input questions into machine-readable reformulations. In ~\cite{qa_rewrite}, Svitlana et al. propose a Transformer Decoder-based model for question rewriting. 
According to a study in ~\cite{reformulation_not_good}, most question reformulation methods  only improve the performance about 2-3\%.  Despite LLMs have exploded in popularity for all sorts of natural language tasks, the ConvQA performance based on LLMs generated reformulations is still upper bounded by human reformulations ~\cite{qa_rewrite, reformulation_not_good}.  

\begin{figure}
	\centering
	\includegraphics[width=0.48\textwidth]{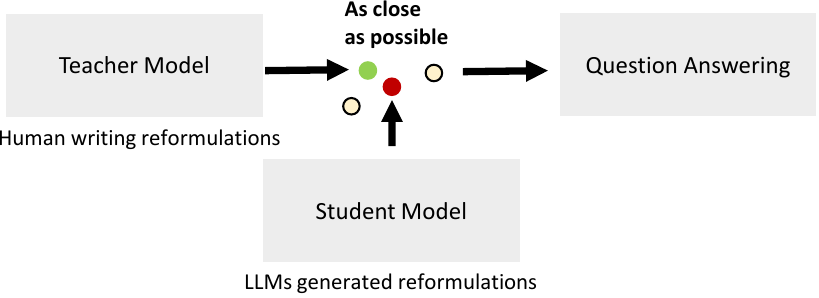}
	\vspace{-1\baselineskip}
	\caption{Reformulation imitator.}
	\label{reformulation_imitation}
	\vspace{-1\baselineskip}
\end{figure}

To further improve the student model's performance, we propose a \textit{teacher-student approach} where a teacher network is trained on human writing reformulations. The teacher network has the same network structure as the student model, but uses human writing reformulations as the input. An example is given in Figure ~\ref{reformulation_imitation}. 
During the training process, our goal is to make the output question embedding of the student model as close as possible to the output of the teacher model in the embedding space. The distance between the student's and teacher's output is measured using the L2 distance:
\begin{equation}\label{query_recommend_loss}
L = \sum_{q_i \in C} [d(\Upsilon_{q_i}, \mathbf{{l}}_{q_i})],
\end{equation}
where $\Upsilon_{q_i}$ is the output of the teacher network for input question $q_i$, and ${l}_{q_i}$ is the output of the student network for the same input with reformulations. By minimizing this distance, we can ensure that the student network is producing output that is similar to that of the teacher model, even when it only has access to the synthetic reformulations. 
Note that the teacher model is pretrained and fixed when we train the student model.  
During the testing phase, given a question, we first use LLMs to generate multiple reformulations for it, then the student model is used to encode the input question with LLMs generated reformulations.  
The performance of directly using the teacher model on the test data is slightly inferior to our model due to the different data distribution of human writing reformulations compared to the reformulations generated with LLMs.

\subsection{Inferring the Topic Entity}


During a conversation, the topic entity may change over time. 
To accurately answer questions, we determine the current topic entity based on the conversation history and the current question. 
We use a multi-layer perception (MLP) to determine whether the topic entity unchanged. If the classifier predicts that the topic entity of the current question is not the answer to the previous question, we set the topic entity to the main topic entity $v_{q_1}$.


The classifier consists of a feed forward neural network (FFN) with ReLU activation functions and a classification layer. The classification layer uses softmax on a 2D output to calculate the cross-entropy loss. Here we use the index format of PyTorch to show that the probability of $v_{q_i} = a^{i-1}$ is equal to the 2nd element of the 2D output. 
\begin{align*}
    \textrm{Pr}(v_{q_i} = a^{i-1}) &= \textrm{MLP\_classifier}(\textrm{FNN}(\mathbf{{l}}_{q_i}))[1]
\end{align*}

\noindent{\bf A - Pretrain Topic Entity Selector.}
The goal of the Topic Entity Selector is to identify the correct topic entity within the conversation, which is an input to the RL model. In order to stabilize the training process for the RL model, we pre-train the parameters of the classifier using binary cross-entropy loss
\begin{align*}
\mathcal{L}_{1} =& -[y\log(\textrm{Pr}(v_{q_i} = a^{i-1})) + \\
& (1 - y)\log(1 - \textrm{Pr}(v_{q_i} = a^{i-1})))] 
\end{align*}

\subsection{Question Answering}

After obtaining the topic entity, the next step is to find the correct entity to answer the user.
We formulate this problem as a Markov decision process (MDP) which is defined by 
a 5-tuple $(S, A, R, P, \gamma)$, 
where $S$ is the state space, $A$ is the action space, $P$ is the state transition function and $R$ denotes the reward function. 

\noindent \textbf{States}. Intuitively, we want a state to encode the question, the current position of the agent in the KG, and the search history information. 
At the $i$th step, the state $s_i \in S$ is defined  as a triple $s_t = (n_i, \mathbf{l}_q, \mathbf{g}_i)$,
$n_i$ is the current entity where the agent is at; $\mathbf{l}_q$ is the question embedding generated by the previous method; and $\mathbf{g}_i$ refers to the search history information.
($n_i, \mathbf{g}_i$) can be viewed as state-dependent information while ($\mathbf{l}_q$) is the global context shared by all states. 

\noindent \textbf{Actions}. The set of possible actions $A_s$ from a state $s_t = (n_t, \mathbf{l}_q, \mathbf{g}_t)$ consists of all outgoing edges of the vertex $n_t$ in the KG.
Formally, $A_s$ = \{($\mathbf{r}_i, \mathbf{u}_{r_i}, \mathbf{e}_{e'}) | (n_t, r_i, e') \in G$\}. 
This means an agent at each state has the option to select which outgoing edge it wishes to take having knowledge of the label of the edge $r_i$ and destination vertex $e'$.
Note that different from most of the existing methods ~\cite{LinRX2018_MultiHopKG, conquer} which only use $\mathbf{A}_s = (\mathbf{r}_i, \mathbf{e}_{e'})$, we also use the unique edge embedding $\mathbf{u}_{r_i}$.
To allow the agent to have the option of ending a search, a self-loop edge is added to every entity. 
In addition, we also include the inverse relationship of a triple in the graph. 

\noindent \textbf{Transition}. The transition function is defined as $\delta : S \times A \longrightarrow S$, which represents the probability distribution of the next
states $\delta(s_{t+1} | s_t, a_t)$. In the current state  $s_t$, the agent aims to choose proper actions $a_t$ and then reach the next state 
$s_{t+1} = (n_{t+1}, \mathbf{l}_q, \mathbf{g}_{t+1})$. 
The $n_t$ and $g_t$ are updated, while the query and answer remains the same.

\noindent \textbf{Rewards}. The model will receive the reward of $R_b(s_t) = 1$ if the current location is the correct answer and 0 otherwise. We set  $\gamma = 1$ during the experiments. 

\subsection{Policy Network}

The search policy is parameterized using state information and global context, including the search history.
Specifically, every entity and relation in $\mathcal{G}$ is assigned a dense vector embedding $\mathbf{e} \in \mathbb{R}^d$ and $\mathbf{r} \in \mathbb{R}^d$ respectively. The action $a_t = (\mathbf{r}_{r_i}, \mathbf{u}_{r_i}, \mathbf{e}_{e'}) \in A_t$ is represented as the concatenation of the relation embedding, the unique edge embedding and the end node embedding.

The search history $(n_1=v_{q_i}, r_1, n_2, ..., n_t) \in H$ consists of the sequence of observations and actions taken up to step $t$, and can be encoded using an LSTM:
\begin{align*}
\mathbf{g}_{0} &= \textrm{LSTM}(0, [\mathbf{e}_{v_{q_i}} || \mathbf{l}_{q_i}]) \\
\mathbf{g}_{t} &= \textrm{LSTM}(\mathbf{g}_{t-1}, \mathbf{a}_{t-1}), t > 0
\end{align*}
where $\mathbf{l}_{q_i}$ is the question embedding to form a start action with $\mathbf{e}_{v_{q_i}}$.
The action space is encoded by stacking the embeddings of all actions in $A_t$: 
$\mathbf{A}_t \in \mathbb{R}^{|A_t|\times 3d}$. And the policy network $\pi$ is defined as:
\begin{align*}
\pi_{\theta}(a_t |s_t) = \delta(\mathbf{A}_t \times \mathbf{W_2} \textrm{ReLU}(\mathbf{W}_1 [\mathbf{n}_t || \mathbf{l}_{q_i} || \mathbf{g}_t]))
\end{align*}
where $\delta$ is the softmax operator. 

\subsection{Knowledge-Based Soft Reward}

Due to the weak supervision in ConvQA, the agent will receive a positive reward until it arrives at the target entity. Such delayed and sparse rewards significantly slow the convergence.
To address the issue of weak supervision and sparsity of rewards in ConvQA, we assign a soft reward to entities other than the target answer to measure the similarity between them. 
This helps to speed up convergence and mitigate incompleteness in the KG. Specifically, the soft reward is used to measure the similarity between the current entity $n_t$ identified by our model and the ground truth answer $a^t$. 
We use ComplEx ~\cite{complEx} to learn the initial entity embedding and relation embedding for all nodes and edges in the knowledge graph. The probability that $n_t$ is the correct answer is calculated by 
\begin{equation}\label{single_probability}
Pr(n_t | \mathbf{l}_{q_i}, v_{q_i}, \mathcal{G}) = Re(<\mathbf{l}_{q_i}, \mathbf{e}_{n_t}, \overline{\mathbf{e}}_{v_{q_i}}>)
\end{equation}

We propose the following soft reward calculating strategy
\begin{align*}
R(s_t) = R_b(s_t) + (1 - R_b(s_t))Pr(n_t | \mathbf{l}_{q_i}, v_{q_i}, \mathcal{G}) 
\end{align*}

Namely, if the destination $n_t$ is a correct answer according to $\mathcal{G}$, the agent receives reward 1. Otherwise the agent receives a fact score weighted by a pretrained distribution: $Pr(n_t | \mathbf{l}_{q_i}, v_{q_i}, \mathcal{G}) $.

\subsection{Training}


Given a set of conversations, we want to return the best possible answers $a^{*}$, maximizing a reward $a^{*} = argmax_a \sum_C \sum_T R(a^i|q_i)$. The reward is computed with respect to the question $q_i$ while the answer is provided in the train dataset. The goal is to maximize the expected reward of the answer returned under the policy $E_{a_1, ..., a_T \sim \pi_{\theta}}[R(s_t)]$. Since it is difficult to compute the expectation, we use Monte Carlo sampling to obtain an unbiased estimate: 
\begin{align*}
E_{a_1, ..., a_T \sim \pi_{\theta}}[R(s_t)] \approx \frac{1}{N} \sum_{i=1}^N \sum_{j=1}^T R(s_t) \pi_{\theta}(a_t|s_t)
\end{align*}
In the experiment, we approximate the expected reward by running multiple rollouts for each training example. The number of rollouts is fixed, We set this number to 20. We use REINFORCE ~\cite{reinforce} to compute gradients for training.
\begin{align*}
&\bigtriangledown_{\theta} E_{a_1, ..., a_T \sim \pi_{\theta}}[R(s_t)] = \sum_{i=1}^T \bigtriangledown_{\theta} \pi_{\theta}(a_t|s_t) R(s_t)  \\
&\approx \frac{1}{N} \sum_{i=1}^N  \sum_{j=1}^T R(s_t) \bigtriangledown_{\theta} \textrm{log}( \pi_{\theta}(a_t|s_t)) \\
\end{align*}
Additionally, to encourage diversity in the paths sampled by the policy during training, we add an entropy regularization term to our cost function, as proposed in ~\cite{conquer}.
\begin{align*}
H_{\pi_{\theta}}(., s) = - \sum_{a \in A_s} \pi_{\theta}(a|s) \textrm{log} \pi_{\theta}(a|s)
\end{align*}
$H_{\pi, \theta}$ is added to the update to ensure better exploration and prevent the agent from getting stuck in local optima.
This final objective is:
\begin{align*}
E_{a_1, ..., a_T \sim \pi_{\theta}}[R(s_t)] + \lambda H_{\pi_{\theta}}(., s) 
\end{align*}

After training, in the testing phase, given a query, we rank all the entities in the KG based on their probability of being the correct answer. We let the policy network keep top-$k$ most likely paths according to beam search, and we rank them according to their possibilities. For all the other entities which are not in the top-$k$ candidates, we use ComplEx ~\cite{complEx} to rank them according to Eq.~\eqref{single_probability}.

\label{sec:method}

\section{Experiments }\label{experiments}

In this section, we evaluate the performance of the proposed \myModel\ algorithm on several public datasets. Our aim is to answer the following questions: (1) How accurate is the proposed \myModel\ algorithm for ConvQA? (2) How efficient is the proposed \myModel\ algorithm?
The code and datasets will be made publically available upon acceptance of the paper. 

\subsection{Experimental Setting}

We use two datasets in the experiments: ConvQuestions ~\cite{convex} and ConvRef ~\cite{conquer}.
ConvQuestions contains a total of 6,720 conversations, each with 5 turns.
ConvRef contains a total of 6,720 conversations, each with 5 turns.
All the conversations in ConvQuestions and ConvRef belong to one of the five domains: "Books", "Movies", "Soccer", "Music", and "TV Series".
Each domain contains 1344 training conversations, 448 validation conversations and 448 test conversations. 
The details of these datasets can be found in Appendix.

Both ConvQuestions and ConvRef use Wikidata~\footnote{\url{https://www.wikidata.org/wiki/Wikidata:Database_download}} as their background KG. However, the full Wikidata KG is extremely large, containing approximately 2 billion triples. Therefore, in the experiment, we sample a subset of triples from Wikidata. We first take the overlapped entities between the Wikidata and the QA datasets, and then we further obtain all the one-hop neighbours of these overlapped entities. The one-hop neighbors are retrieved from both the original data dump and also the entities' corresponding online Wikidata websites. 

\hide{
\begin{table}[]
	\caption{ Hyperparameters for \myModel. }
	\begin{tabular}{|c|c|}
		\hline
		{\bf Hyperparameters}       & {\bf Value}                \\ \hline
        \multicolumn{2}{|c|}{ComplEx}     \\ \hline
		epochs         & 100 \\ 
		batch size         & 256 \\ 
		learning rate       & 0.005 \\
		embedding dimension         & 200 \\ 
        dropout ratio &  0.2 \\ 
        optimizer &     Adam   \\ \hline
        \multicolumn{2}{|c|}{\myModel}     \\ \hline
		epochs         & 20 \\ 
		batch size         & 16 \\ 
		learning rate       & 0.00002 \\
        rollout number & 20 \\
		embedding dimension of policy network        & 600 \\ 
        optimizer &     Adam   \\ \hline
	\end{tabular}
\label{Hyperparameters}
\end{table}
}

We compare the performance of our method, \myModel, with four baselines:
\textbf{Convex} ~\cite{convex}: this method detects answers to conversational utterances over KGs by first extracting a subgraph, then identifying answers in the subgraph. 
\textbf{Conqer} ~\cite{conquer}: this is the current state-of-the-art baseline. It uses RL with reformulations to find answers in the KG. 
\textbf{OAT} ~\cite{OAT}: this Transformer-based model takes a JSON-like structure as input to generate a Logical Form (LF) grammar that can model a wide range of queries on the graph. It finds answers by applying the LF.
\textbf{Focal Entity} ~\cite{lan-jiang-2021-modeling}: this is a novel graph-based model to find answer by graph neural network.

Two LLMs are used to generate reformulations for the input query, which are GPT2 ~\cite{gpt2} and Bart ~\cite{bart}. For each input question, we generate multiple reformulations and use attention mechanism to aggregate them inside the model. 
We adopt the following ranking metrics which are also employed by the previous baselines: (1) Precision at the top rank (P@1); (2) Mean Reciprocal Rank (MRR) is the average across the reciprocal of the rank at which the first context path was retrieved; (3) Hit ratio at k (H@k/Hit@k) is the fraction of times a correct answer was retrieved within the top-k positions. 
The details of all the datasets and experiment environment can be found in Appendix.

\begin{table*}
	\centering
	\caption{Performance on different domain datasets.}
 \vspace{-0.5\baselineskip}
    \scalebox{0.8}{
	\begin{tabular}{|c|c|c|c|c|c|c|c|c|c|c|c|c|c|c|c|c|c|}
	\hline
    Domain  & \multicolumn{3}{c|}{Movies} & \multicolumn{3}{c|}{TV Series}  & \multicolumn{3}{c|}{Music}   & \multicolumn{3}{c|}{Soccer} \\ \hline
    Metric & H@3  &  H@5  & H@8 & H@3  &  H@5  & H@8  & H@3  &  H@5  & H@8  & H@3  &  H@5  & H@8  \\ \hline
	\multicolumn{1}{|c|}{Dataset} & \multicolumn{12}{c|}{ConvQA}   \\ \hline
	Conv & 0.345 & 0.345 & 0.345 & 0.303 & 0.308 & 0.309 & 0.213 & 0.217 & 0.221 &  0.2019 & 0.2019 & 0.2019 \\ \hline
	Conquer & 0.336 & 0.343 & 0.354 & 0.375 & 0.396 & 0.421 & 0.279 & 0.282 & 0.289  & \cellcolor{lightgray} 0.325 & 0.343 & 0.350 \\ \hline
    \rowcolor{lightgray} \myModel\ & 0.385 & 0.456 & 0.514 & 0.437 & 0.482 & 0.587 & 0.290 &  0.356 & 0.392 &  \cellcolor{white} 0.288 & 0.418 &  0.550 \\ \hline
    \multicolumn{1}{|c|}{Dataset} & \multicolumn{12}{c|}{ConvRef}   \\ \hline
	Conv & 0.345 & 0.345 & 0.345 & 0.303 & 0.308 & 0.308 & 0.213 & 0.217 &  0.221 &  0.202 & 0.202 & 0.202 \\ \hline
	Conquer & 0.389  & 0.404 &  0.429 & 0.435 & 0.442 & 0.485 & 0.371 & 0.398 & 0.413  & \cellcolor{lightgray} 0.371 & 0.384 & 0.393 \\ \hline
\rowcolor{lightgray}	\myModel\ &  0.393 & 0.461 & 0.521 & 0.436 & 0.533 & 0.564 & 0.377 & 0.447 & 0.484 & \cellcolor{white} 0.353 & 0.385 & 0.425 \\ \hline
	\end{tabular}
 }
	\label{domain}
 \vspace{-1\baselineskip}
\end{table*}

\subsection{Main Results}

In this subsection, we test \myModel\ on conversational question answering tasks and compare it with other baseline methods. 

\noindent{\bf A - Overall performance on ConvQA datasets.}
Table ~\ref{overall} compares the results of \myModel\ with baselines on the ConvQuestions and ConvRef datasets. 
As we can see, \myModel\ outperforms the previous baselines in the H@5 and MRR metrics On ConvQA.
For H@5, \myModel\ performs 4.5\% better than CONQUER and 20\% better than CONVEX.
In terms of MRR, CONVEX has the lowest performance, which is 13.7\% worse than \myModel. 
CONQUER has the second highest performance, but it is also 1\% lower than \myModel.
For OAT, because its source code is not available, we directly adopt its results from ~\cite{OAT}.
We can find that it has the highest P@1 compared to other methods. However, its MRR is 7.3\% points lower than that of \myModel.
For Focal Entity, it has the second highest P@1 and the third highest MRR.
On the ConvRef dataset, \myModel\ also has similar performance. It achieves the highest Hit@5, which is 5\% better than that of CONQUER. 
\myModel\ also has the second highest P@1 and MRR compared with other baselines. 
Due to the unavailability of the OAT source code and the failure to run Focal Entity on the ConvRef dataset, we are unable to include their results in our analysis of the ConvRef dataset.


\begin{table}
	\centering
	\caption{Performance on ConvQA and ConvRef.}
	\vspace{-0.5\baselineskip}
 \scalebox{0.8}{
	\begin{tabular}{|c|c|c|c|c|c|c|}
	\hline
	\multicolumn{1}{|c|}{Dataset} & \multicolumn{2}{c|}{ConvQA}  &  \multicolumn{2}{c|}{ConvRef}   \\ \hline
	Model        & Hit@5 & MRR  & Hit@5 & MRR  \\ \hline
	CONVEX   &  0.219  & 0.200   & 0.257 & 0.241 \\ \hline
	CONQUER   &  0.372   & 0.327  & 0.427 & \cellcolor{lightgray} 0.382 \\ \hline
	OAT  & -   &  0.260   & -  & - \\ \hline
	Focal Entity  & -  & 0.248 & -  & - \\ \hline
\rowcolor{lightgray}	\myModel\  & 0.417  & 0.337  & 0.477 & \cellcolor{white} 0.353 \\ \hline
	\end{tabular}}
	\label{overall}
	\vspace{-1\baselineskip}
\end{table}


\noindent{\bf B - Performance on different domains.}
We further investigate the ranking performance of \myModel\ across different domains for both benchmarks. Table ~\ref{domain} illustrates detailed ranking results for H@3, H@5 and H@8. 
As the results show, \myModel\ outperforms other baselines on most domains in the ConvQuestions benchmark. On average, it achieves a 13.7\% improvement in H@8 and 6.6\% improvement in H@5 compared to the second highest baseline CONQUER. It performs slightly worse on H@3, with an average of 0.58\% lower than CONQUER. The relatively poor performance on the Books domain is likely due to the presence of ``yes/no" questions and queries regarding the plot, making it difficult for the topic entity selector to accurately determine the topic entity, posing a challenge for the model's predictions.
Similar results are also observed on the ConvRef dataset.

\subsection{Ablation Studies and Efficiency Results} In this subsection, we show the effectiveness of by ablation Studies.

\noindent{\bf A - The effectiveness of the reformulations.}
In this subsection, we demonstrate the effectiveness of using different reformulations in the model. 
Two large language models are used to generate reformulations: GPT2 and Bart.
When not using reformulations, the output of the Question Encoder is treated as the input of the LSTM directly. Table ~\ref{reformulations} shows the experiment results.
As we can see, using reformulations can indeed increase the ConvQA performance most of the time. The performance of using GPT2 reformulations is very similar to that of using Bart reformulations. Using human writing reformulations has the best performance. 


\begin{table}[H]
	\caption{The effectiveness of the reformulations.}
	\vspace{-1\baselineskip}
 \scalebox{0.8}{
	\begin{tabular}{|c|c|c|c|c|c|c|}
	\hline
	Model          & P@1 & Hit@3 & Hit@5  & Hit@8 \hide{& MRR}   \\ \hline
	No Reformulation & 0.231   &   0.373  & 0.445  & 0.474 \hide{& 0.327} \\ \hline
    GPT2 Reformulation & 0.212  & 0.367 &  0.451 & 0.504  \hide{& 0.313} \\ \hline
    Bart Reformulation & 0.221  & 0.376 &  0.441 & 0.494  \hide{& 0.320} \\ \hline
    \cellcolor{lightgray} Human  Reformulation  & \cellcolor{lightgray} 0.257  & \cellcolor{lightgray} 0.395 &  \cellcolor{lightgray} 0.463 & \cellcolor{lightgray}0.518  \hide{& 0.346} \\ \hline
	\end{tabular}
 }
	\label{reformulations}
	\vspace{-1\baselineskip}
\end{table}

We further test the effectiveness of the proposed teacher-student model, shown in Table ~\ref{teacher-student}. If we only use the reformulations generated by LLMs, the performance is about 3\% lower than that of teacher-student model. If we train the model on human writing reformulations while tests on generated reformulations, the performance is about 1.3\% lower than \myModel. 

\begin{table}[H]
	\caption{The effectiveness of the teacher-student model.}
	\vspace{-1\baselineskip}
 \scalebox{0.8}{
	\begin{tabular}{|c|c|c|c|c|c|c|}
	\hline
	Model          & P@1 & Hit@3 & Hit@5  & Hit@8 \hide{& MRR}   \\ \hline
 GPT2 Reformulation & 0.212  & 0.367 &  0.451 & 0.504  \hide{& 0.313} \\ \hline
 \rowcolor{lightgray} \myModel\ (GPT2)  & 0.265  & 0.404 &  0.477 & 0.526  \hide{& 0.313} \\ \hline
 Bart Reformulation & 0.221  & 0.376 &  0.441 & 0.494  \hide{& 0.320} \\ \hline
 \rowcolor{lightgray} \myModel\ (Bart)  & 0.244  & 0.413 &  0.491 & 0.523  \hide{& 0.3125} \\ \hline
 Train test different & 0.237  & 0.389 &  0.472 & 0.507  \hide{& 0.320} \\ \hline
	\end{tabular}}
	\label{teacher-student}
	\vspace{-1\baselineskip}
\end{table}

\begin{figure}[H]
	\centering
	\includegraphics[width=0.4\textwidth]{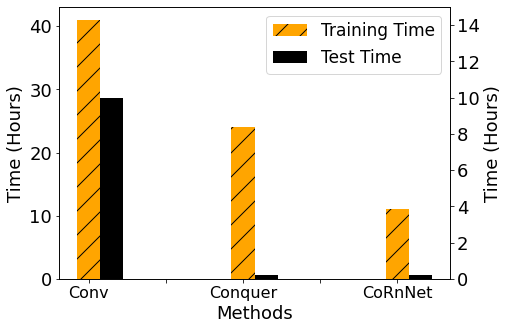}
	\vspace{-1\baselineskip}
	\caption{\myModel\ Training and Test Time.}
	\label{train_test_time}
 \vspace{-1\baselineskip}
\end{figure}

\noindent{\bf C - Efficiency.}
Figure ~\ref{train_test_time} shows the training time and test time of different methods on ConvRef dataset. As we can see, \myModel\ has the shortest training and test time. While Conv has the longest training and test time. 
Despite the short training time of \myModel, it can still achieve better or comparable ConvQA performance compared to other baselines.

\hide{
\noindent{\bf E - A successful case of using reformulation.}
Here we present a successful instance demonstrating the efficacy of reformulation in improving comprehension. Regarding question Q3, its lack of explicitness makes it challenging for most methods to infer the user's intent to inquire about 'where was Dan Brown born?' However, leveraging reformulation facilitates the model in swiftly identifying the correct answer. 

\begin{myquote}

Q1: Who is the author of the book Inferno?

A1: Dan Brown

Q2: And when was the author born?

A1: 1964-06-22T00:00:00Z

Q3: And where?
\begin{inside_myquote}
Without reformulations, the method gives the wrong answer. 
\end{inside_myquote}
\begin{inside_myquote}
With reformulations "In which city was he born?", the model can find the correct answer "Exeter".
\end{inside_myquote}
\end{myquote}

}

\section{Related work}\label{related-work}

\subsection{Conversational Question Answering}
Various approaches have been used to develop ConvQA systems. 
For instance, in ~\cite{ask_the_right_question}, the authors employed RL to train an agent that reformulates input questions to aid the system's understanding. In ~\cite{Dialog-to-Action}, an encoder-decoder model is used to transform natural language questions into logical queries for finding answers. In ~\cite{kacupaj-etal-2021-conversational}, a Transformer model is used to generate logical forms and graph attention is introduced to identify entities in the query context. 
Other systems, such as Google's Lambda~\cite{google_lamda}, Apple's Siri, and OpenAI's ChatGPT, are also pursuing this task.

\subsection{Knowledge Graph Question Answering}
While knowledge graph question answering has been researched for some time, many of the existing methods primarily focus on answering single-turn questions ~\cite{binet, prefnet} or complex questions ~\cite{wang2021benchmarking, gfinder}. For example, Zhang et al ~\cite{ZHANG2022102933} use a KG as the environment and propose a RL-based agent model to navigate the KG in order to find answers to input questions. Similarly, in ~\cite{go_for_a_walk, LinRX2018_MultiHopKG, deeppath, adam}, authors use RL models to find paths in the KG for answering input queries.
Other studies, such as ~\cite{misu-etal-2012-reinforcement, xiaobing, alexa, gpt2, gpt3}, integrated RL with other methods to create more human-like systems.
Some other works try to use RL to tackle multi-turn conversations. For example, in ~\cite{dhingra-etal-2017-towards}, the authors proposed a multi-turn dialogue agent which helps users search Knowledge Bases (KBs) without composing complicated queries. 
In ~\cite{ZHANG2022102933}, instead of using a random walk agent, an adaptive path generator is developed with several atomic operations to sequentially generate the relation paths until the agent reaches the target entity. 
However, only a few of these studies have attempted to utilize reformulations to enhance KGQA performance, as opposed to the focus of this paper. 

\subsection{Question Rewriting}
Question Rewriting 
aims to reformulate an input question into a more salient representation. This can improve the accuracy of search engine results or make a question more understandable for a natural language processing (NLP) system.
In ~\cite{qa_rewrite}, an unidirectional Transformer decoder is proposed to automatically rewrite a user's input question to improve the performance of a conversational question answering system. 
In ~\cite{elgohary-etal-2019-unpack}, authors proposed a Seq2Seq model to rewrite the current question according to the conversational history, and also introduced a new dataset named CANARD.
In ~\cite{10.1145/2623330.2623677}, query rewrite rules are mined from a background KG and a query rewrite operator is used to generate a new question.

Unlike the previous techniques, \myModel\ trains teacher-student model with both human writing reformulations and LLMs generated reformulations. This approach helps to avoid the negatively impact from the generated low quality reformulations.

\section{Conclusion}\label{conclusion}

In this paper, a model (\myModel) that creatively combines the question reformulation and reinforcement learning is proposed on a knowledge graph (KG) to attain accurate multi-turn conversational question answering. \myModel\ utilizes a teacher-student distillation approach and reinforcement learning to find answers from a KG. 
Experimental results demonstrate that \myModel\ surpasses existing methods on various benchmark datasets on conversational question answering.

\bibliographystyle{plplainurl}
\bibliography{008reference.bib}

\begin{thebibliography}{10}

\bibitem{google_lamda}
LaMDA: Language Models for Dialog Applications, 2022.
\newblock Adres: \url{https://arxiv.org/abs/2201.08239}.

\bibitem{alexa}
A~Acharya,  S~Adhikari.
\newblock Alexa Conversations: An Extensible Data-driven Approach for Building Task-oriented Dialogue Systems.
\newblock 2021.

\bibitem{gpt3}
T~Brown, B~Mann, N~Ryder.
\newblock Language Models are Few-Shot Learners.
\newblock {\em Advances in Neural Information Processing Systems}.

\bibitem{ask_the_right_question}
C~Buck,  J~Bulian.
\newblock Ask the Right Questions: Active Question Reformulation with Reinforcement Learning.
\newblock 2017.

\bibitem{convex}
P~Christmann, Saha R, A~Abujabal, J~Singh, G~Weikum.
\newblock Look before You Hop: Conversational Question Answering over Knowledge Graphs Using Judicious Context Expansion.
\newblock CIKM '19, 2019.

\bibitem{go_for_a_walk}
R~Das, S~Dhuliawala, M~Zaheer.
\newblock Go for a Walk and Arrive at the Answer: Reasoning Over Paths in Knowledge Bases using Reinforcement Learning, 2017.

\bibitem{bert}
Jacob Devlin, Ming-Wei Chang, Kenton Lee, Kristina Toutanova.
\newblock BERT: Pre-training of Deep Bidirectional Transformers for Language Understanding, 2019.

\bibitem{dhingra-etal-2017-towards}
B~Dhingra, L~Li, X~Li, J~Gao, Li~Deng.
\newblock Towards End-to-End Reinforcement Learning of Dialogue Agents for Information Access.
\newblock Association for Computational Linguistics.

\bibitem{elgohary-etal-2019-unpack}
A~Elgohary, D~Peskov, J~Boyd-Graber.
\newblock Can You Unpack That? Learning to Rewrite Questions-in-Context.
\newblock Association for Computational Linguistics, 2019.

\bibitem{10.1145/2623330.2623677}
A~Fader, L~Zettlemoyer, O~Etzioni.
\newblock Open Question Answering over Curated and Extracted Knowledge Bases.
\newblock KDD '14. Association for Computing Machinery, 2014.

\bibitem{Dialog-to-Action}
D~Guo, D~Tang, N~Duan, M~Zhou, J~Yin.
\newblock Dialog-to-Action: Conversational Question Answering Over a Large-Scale Knowledge Base.
\newblock Curran Associates, Inc., 2018.

\bibitem{reformulation_not_good}
Etsuko Ishii, Yan Xu, Samuel Cahyawijaya, Bryan Wilie.
\newblock Can Question Rewriting Help Conversational Question Answering?, 2022.

\bibitem{kacupaj-etal-2021-conversational}
E~Kacupaj, J~Plepi, K~Singh, H~Thakkar.
\newblock Conversational Question Answering over Knowledge Graphs with Transformer and Graph Attention Networks.
\newblock Association for Computational Linguistics.

\bibitem{contrastiveQA}
E~Kacupaj, K~Singh, M~Maleshkova, J~Lehmann.
\newblock Contrastive Representation Learning for Conversational Question Answering over Knowledge Graphs, 2022.

\bibitem{conquer}
M~Kaiser, R~Saha~Roy, G~Weikum.
\newblock Reinforcement Learning from Reformulations in Conversational Question Answering over Knowledge Graphs.
\newblock SIGIR '21. Association for Computing Machinery, 2021.

\bibitem{adam}
Diederik~P. Kingma,  Jimmy Ba.
\newblock Adam: A Method for Stochastic Optimization, 2017.

\bibitem{lan-jiang-2021-modeling}
Y~Lan,  J~Jiang.
\newblock Modeling Transitions of Focal Entities for Conversational Knowledge Base Question Answering.
\newblock Association for Computational Linguistics, 2021.

\bibitem{bart}
M~Lewis, Y~Liu, N~Goyal.
\newblock BART: Denoising Sequence-to-Sequence Pre-training for Natural Language Generation, Translation, and Comprehension, 2019.

\bibitem{LinRX2018_MultiHopKG}
X~Lin,  R~Socher.
\newblock Multi-Hop Knowledge Graph Reasoning with Reward Shaping.
\newblock {\em {EMNLP} 2018}.

\bibitem{gfinder}
L.~{Liu}, B.~{Du}, J.~{xu}, H.~{Tong}.
\newblock G-Finder: Approximate Attributed Subgraph Matching.
\newblock {\em 2019 IEEE International Conference on Big Data (Big Data)}, strony 513--522, 2019.

\bibitem{prefnet}
Lihui Liu, Yuzhong Chen, Mahashweta Das, Hao Yang, Hanghang Tong.
\newblock Knowledge Graph Question Answering with Ambiguous Query.
\newblock {\em Proceedings of the ACM Web Conference 2023}, 2023.

\bibitem{binet}
Lihui Liu, Boxin Du, Jiejun Xu, Yinglong Xia, Hanghang Tong.
\newblock Joint Knowledge Graph Completion and Question Answering.
\newblock {\em Proceedings of the 28th ACM SIGKDD Conference on Knowledge Discovery and Data Mining}, 2022.

\bibitem{OAT}
P~Marion, K~Nowak, F~Piccinno.
\newblock Structured Context and High-Coverage Grammar for Conversational Question Answering over Knowledge Graphs, 2021.

\bibitem{misu-etal-2012-reinforcement}
T~Misu, K~Georgila, A~Leuski, D~Traum.
\newblock Reinforcement Learning of Question-Answering Dialogue Policies for Virtual Museum Guides.
\newblock Association for Computational Linguistics, 2012.

\bibitem{SL-Oracle}
R~Nogueira,  K~Cho.
\newblock Task-Oriented Query Reformulation with Reinforcement Learning, 2017.

\bibitem{gpt2}
A~Radford, J~Wu, R~Child.
\newblock Language Models are Unsupervised Multitask Learners.
\newblock 2018.

\bibitem{complEx}
T~Trouillon, J~Welbl, S~Riedel.
\newblock Complex Embeddings for Simple Link Prediction.
\newblock ICML'16. JMLR.org, 2016.

\bibitem{qa_rewrite}
S~Vakulenko, S~Longpre, Z~Tu, R~Anantha.
\newblock Question Rewriting for Conversational Question Answering.
\newblock WSDM '21. Association for Computing Machinery.

\bibitem{transformer}
Ashish Vaswani, Noam Shazeer, Niki Parmar, Jakob Uszkoreit, Llion Jones, Aidan~N. Gomez, Lukasz Kaiser, Illia Polosukhin.
\newblock Attention Is All You Need, 2017.

\bibitem{wang2021benchmarking}
Zihao Wang, Hang Yin, Yangqiu Song.
\newblock Benchmarking the combinatorial generalizability of complex query answering on knowledge graphs.
\newblock {\em Proceedings of the 35th International Conference on Neural Information Processing Systems}, 2021.

\bibitem{reinforce}
Ronald~J. Williams.
\newblock Simple Statistical Gradient-Following Algorithms for Connectionist Reinforcement Learning.
\newblock {\em Mach. Learn.}, 1992.

\bibitem{deeppath}
Wenhan Xiong, Thien Hoang, William~Yang Wang.
\newblock DeepPath: A Reinforcement Learning Method for Knowledge Graph Reasoning, 2017.
\newblock Adres: \url{https://arxiv.org/abs/1707.06690}.

\bibitem{ZHANG2022102933}
Qixuan Zhang, Xinyi Weng, Guangyou Zhou, Yi~Zhang, Jimmy~Xiangji Huang.
\newblock ARL: An adaptive reinforcement learning framework for complex question answering over knowledge base.
\newblock {\em Information Processing and Management}, 59(3):102933, 2022.
\newblock Adres: \url{https://www.sciencedirect.com/science/article/pii/S0306457322000565}.

\bibitem{xiaobing}
Li~Zhou, Jianfeng Gao, Di~Li, Heung-Yeung Shum.
\newblock {The Design and Implementation of XiaoIce, an Empathetic Social Chatbot}.
\newblock {\em Computational Linguistics}, 46(1):53--93, 03 2020.
\newblock Adres: \url{https://doi.org/10.1162/coli\_a\_00368}.

\end{thebibliography}


\end{document}